# Why Dyslexic Reading Takes Longer – and When: Quantifying the Effects of Word Length, Frequency, and Predictability on Dyslexia

Hugo Rydel- Johnston[1] and Alexandros Kafkas[1]

[1]Division of Psychology, Communication & Human Neuroscience, The University of Manchester

## Author Note

Hugo Rydel – Johnston https://orcid.org/0009-0006-1103-1015

Alex Kafkas https://orcid.org/0000-0001-5133-8827

We have no conflicts of interest to disclose.

Correspondence concerning this article should be addressed to Hugo Rydel-Johnston, Division of Psychology, Communication & Human Neuroscience, The University of Manchester, Oxford Road, Manchester, M13 9PL, UK. Email: hugo.rydel-johnston@student.manchester.ac.uk

**Abstract**

We ask where, and under what conditions, dyslexic reading costs arise in a large-scale naturalistic reading dataset. Using eye-tracking aligned to word-level properties—word length, frequency, and predictability—we model the influence of each of these features on dyslexic time costs. We find that all three properties robustly change reading times in both typical and dyslexic readers, but dyslexic readers show stronger sensitivities to each of the three features, especially predictability. Counterfactual manipulations of these features substantially narrow the dyslexic–control gap—by about one-third—with predictability showing the strongest effect, followed by length, and frequency. These patterns align with existing dyslexia theories suggesting heightened demands on linguistic working memory and phonological encoding in dyslexic reading and directly motivate further research into lexical complexity and preview benefits to further explain the quantified gap. In effect, these findings break down when extra dyslexic costs arise, how large they are, and provide actionable guidance for the development of interventions and computational models for dyslexic readers.

**Why Dyslexic Reading Takes Longer – And When**

Dyslexia is characterized by persistent difficulty in accurate and/or fluent word recognition and decoding (Lyon et al., 2003) and affects between 4–8% of individuals (Yang et al., 2022; Doust et al., 2022). A robust behavioural signature of this condition, observable well into adolescence and adulthood, is a disproportionate slowdown during reading. Dyslexic readers make more eye fixations, those fixations last longer, and they skip words less often than typically developing peers (Hyönä & Olson, 1995; Hutzler & Wimmer, 2004; De Luca et al., 2002). This global pattern has been documented across orthographies and age groups, in large reviews and targeted studies, using connected text, word- and pseudoword lists (Rayner, 1998; Hutzler & Wimmer, 2004; Hawelka et al., 2010). Yet, despite consensus on the existence of a dyslexia slowdown, we still lack process-linked, interpretable markers that explain why these time costs occur, when in processing they arise, and how much they contribute to the dyslexic–control gap in reading times. The core aim of this study is thus to establish a transparent, word-level account of where this time accumulates for dyslexic readers.

During sentence reading, the eyes alternate between brief pauses (fixations) and rapid jumps (saccades). In cognitive research, these two basic measures are often further broken down into skipping, first-fixation duration, and total (per word) reading time, each of which indexes a different cognitive process. Skipping (the binary decision to fixate a word or not) indexes very early signs of ease during word identification. If a person consistently skips words, this reflects successful visual preview and/or strong contextual expectation about upcoming word identities (Schotter, Angele, & Rayner, 2012; Rayner & Well, 1996) – both signs of an experienced reader and/or simpler text (Rayner, 1998; Slattery et al., 2018). Appropriately, skipping rates typically rise when words are short, frequent, and predictable (Rayner et al., 2011; Schotter, Angele, & Rayner, 2012) as well as for more experienced, and older readers (Rayner, 1998; Reichle et al., 2013; Slattery et al., 2018).

First-fixation duration (FFD), the second eye-measure, refers to the earliest foveal processing on an encountered word and is commonly interpreted as tracking initial lexical access to a word (Staub, 2015; Reichle, Rayner, & Pollatsek, 2003). Thus, longer FFDs on a word can index increased difficulties with initial word processing: either due to difficulties identifying the word or integrating it with its prior context (Clifton, Staub, & Rayner, 2007; Staub, 2015). In typical adult readers, FFDs reliably shorten for high-frequency and highly predictable words, reflecting facilitation at the initial lexical-access stage enabled by these word features (Rayner, 1998; Rayner et al., 2011; Staub, 2015).

Finally, total reading time (TRT) sums all fixations on a word, including any second-pass re-readings. While FFD and TRT are similar, increases in TRT also indicate extended processing—such as integration demands or reanalysis (Kliegl et al., 2004; Clifton, Staub, & Rayner, 2007; Staub, 2015). This means TRT is more sensitive to the overall speed of word and context processing of a reader than FFD. TRT is especially sensitive to integration and reanalysis costs that increase with syntactic-memory load (Gibson, 1998; Boston et al., 2008), making it a useful index of later-stage "integration" processes throughout reading.

Taken together, skipping, FFD, and TRT provide a clear division in reading mechanisms: skipping asks *whether* a word needs foveal time at all (measuring preview and identification success), while FFD and TRT ask *how much* time is required for processing a word in context (focusing more on word identification and integration demand measures, respectively).

Building on these distinct measures, three word-level properties moderate whether a word is fixated at all (skipped) and, if fixated, how long early (FFD) and extended (TRT) processing takes. These are word length, lexical frequency, and contextual predictability. The first of these, word length, affects reading times by directly changing how much of a word falls within the high-acuity vision of the fovea; the longer a word is, the more of it (necessarily) spills into parafoveal regions where visual crowding processes make letter identification less reliable and word identity less certain (Schotter & Leinenger, 2016; Pelli & Tillman, 2008 Juhasz et al., 2008). This reduces the odds of one-shot recognition (processing a word using just one fixation) and increases the need for serial (letter-by-letter) fixation to recognize the word (Pelli & Tillman, 2008), both of which increase reading times. Empirical research demonstrates these effects by linking increased word length with decreases in skipping rates, and increased FFD and TRTs (Kliegl et al., 2004; Rayner et al., 2011; Tiffin-Richards & Schroeder, 2015), with effects that remain even after controlling for other factors like frequency and predictability (Kliegl et al., 2004; Rayner et al., 2011).

Lexical frequency, our second moderator, quantifies how common a word is in a language (and thus, how likely a reader is to know and recognize it). Frequent words commonly have more entrenched lexical representations (Perfetti, 2007) (i.e. pathways towards the retrieval of key information about the word), so mapping from initial viewing through to retrieval of the word's phonological and semantic code is faster and more reliable. Within computational accounts of eye-movement control, frequency most strongly affects the "familiarity check" stage (the stage at which you decide if you know a word), predicting shorter first fixations (lower FFD) and reducing the need for extensive word processing (lower TRT) for frequent words (Reichle, Rayner, & Pollatsek, 2003). Empirically, higher frequency words reflect this, increasing skipping and shortening FFD (Rayner, 1998; Pollatsek et al., 2008); when fixated, frequent words also reduce late measures such as TRT, by imposing fewer integration challenges requiring regressions back to the word (Rayner et al., 2004; Tiffin-Richards & Schroeder, 2015).

Contextual predictability, our third eye-measure moderator, captures how expected a word is given its preceding context (Staub, 2011, 2015). Typically, when the next word aligns with the reader's predictions (e.g., *peanut butter and ___ → jelly*), prior context naturally pre-activates lexical and phonological features for the word (Kutas & Federmeier, 2011). For highly predictable words, this means reductions in FFD and TRT and raised skipping rates, as words require less evidence to be correctly identified (Rayner & Well, 1996; Schotter, Angele, & Rayner, 2012; White et al., 2005). By contrast, when a word is unexpected, more foveal time is needed and misanalysis/integration difficulties can cascade to produce longer TRT when the word's context needs to be revisited (Clifton, Staub, & Rayner, 2007; Warren, McConnell, & Rayner, 2008).

Thus, length primarily reduces skipping and lengthens both early (FFD) and late (TRT) measures via parafoveal constraints; frequency facilitates early lexical access (shorter FFD, more skipping) through increasing entrenched lexical representations; and predictability supports both early and late processing (shorter FFD/TRT, more skipping) by pre-activating likely word candidates. Together, these properties jointly determine whether a word is fixated at all and, if so, how much time is spent on initial identification versus later integration/reanalysis.

Dyslexic slowdowns are likely amplified at each processing stage targeted by word length, frequency, and predictability. First, according to the phonological-decoding account, dyslexia is characterized by inefficient mappings from written text to sounds (Ziegler & Goswami, 2005; Peterson & Pennington, 2015), with specific outcomes being slower, less reliable access to phonological codes and corresponding lexical entries amongst dyslexics. This bottleneck likely delays the "familiarity check" that triggers swift saccades to upcoming words, and forces more foveal evidence to be accrued before the familiarity threshold is reached (Reichle, Rayner, & Pollatsek, 2003), producing reading costs at FFD and TRT. The reading cost this incurs also scales as phonological assembly or lexical activation becomes more demanding, such as in long and rare words: long words require more text → sound conversions while low-frequency words have naturally weaker, slower-to-activate lexical and phonological representations for each word. Accordingly, the phonological-decoding account broadly predicts disproportionately long slowdowns while reading (longer FFD) and more, serial, fixations across words (longer TRT) for dyslexics to increase confidence of word identification (Perfetti, 2007). Uncertainty at identification can also spill over into additional TRT costs when initial misinterpretation must be resolved later in the text (Clifton, Staub, & Rayner, 2007). In line with this, dyslexic readers exhibit more and longer fixations as length and frequency increase, with fewer skips, more serial fixations, and longer gaze durations relative to controls (Hyönä & Olson, 1995; Hutzler & Wimmer, 2004; De Luca et al., 2002).

Separately, well-documented working-memory limitations in dyslexia—spanning phonological storage and central-executive control (Maehler & Schuchardt, 2016; Du & Zhang, 2023; Smith-Spark & Fisk, 2007)—may selectively affect the use of context for prediction and thus magnify costs when contextual predictability is low. Even in typical readers, lower predictability reduces skipping and lengthens FFD and TRT (Rayner & Well, 1996; Staub, 2015). Thus, if dyslexic readers retain less context or update predictions less effectively, they should show disproportionately longer fixations and TRT for unexpected words, beyond typical readers. Converging evidence is consistent with this view: adults with dyslexia show attenuated linguistic prediction effects (Engelhardt et al., 2021), and speed-impaired readers (a group encompassing many dyslexics) generate forward inferences to a reduced extent during sentence reading (Hawelka et al., 2015) – both signs of difficulties in next-word prediction. Even so, dyslexia-specific tests of predictability in naturalistic reading are rare, leaving this effect, so far, unquantified.

Taken together, both mechanisms—slowed phonological coding and constrained WM for context—map naturally onto the three word-level features mentioned earlier: length and frequency primarily tax early identification (skipping/FFD/TRT), whereas predictability additionally stresses context maintenance and reanalysis (FFD/TRT), predicting larger time costs for dyslexic readers.

However, despite strong individual literatures on length, frequency, and predictability in typical readers, current evidence falls short of a transparent, word-aligned account of dyslexia, or a dyslexia–control time gap. Additionally, although modelling at the word level is now standard, few studies adopt time-decomposition approaches that report interpretable millisecond shifts and quantify how much of the group gap is explained by each studied factor. This methodological gap matters for theory. Most mainstream models of reading (e.g., E-Z Reader and SWIFT) make *quantitative* predictions about how word-level differences determine eye measure effects and total reading time (Reichle et al., 2003; Engbert et al., 2005; Reichle et al., 2009). Thus, without a word-aligned variance partition we cannot evaluate the relative importance textual features play, and consequently, are unable to contrast empirical findings against these models. This also carries importance for potential applications. Analyses that pair millisecond effect sizes with cross-validated prediction models allow for transparent communication of how much each modifiable factor contributes to a reader's total reading time and enables extension of findings to different languages and contexts with little additional effort for the experimenter, which can support the creation of new screening thresholds, targeted interventions, and research tied to the direct modification of these lexical properties.

The present study addresses these gaps by combining naturalistic eye-tracking during connected-text reading with token-level alignment of word features and eye measures. We first estimate interpretable millisecond shifts associated with word length, lexical frequency, and predictability. Second, we evaluate the relative sensitivity of controls vs. dyslexics to changes in these features. Finally, we quantify how much of the dyslexia–control reading time gap these features jointly explain. Throughout the study, we break down our measure into three key components: skipping rate, total reading time on a word (TRT), and a novel expected reading time (ERT) measure, used to join skipping rate and TRT per word to quantify the *total* contribution a feature has on reading times. We define ERT as:

$$\mathrm{ERT} = \left(1 - P(\mathrm{skip})\right) \times \mathrm{TRT}_{\mathrm{fixated}}$$

To offer clear, quantifiable effect sizes, we also employ a Q1→Q3 ms-shift approach. This means we change our features from their 25th percentile (1st quartile) to the 75th percentile (3rd quartile) to measure millisecond changes in our key measures (skipping, TRT, ERT). With that considered, we outline three main, and one exploratory, hypotheses:

1. Feature effects: A Q1→Q3 shift in each of our three word-level features will significantly change expected reading times (ERT) in the following direction:
   - Word Length ↑ ⇒ ERT ↑.
   - Word Frequency ↑ ⇒ ERT ↓.
   - Word Predictability ↑ (-surprisal) ⇒ ERT ↓.
2. Dyslexic amplification: Dyslexic readers will show amplified (directionally consistent) ERT costs for each of these features.
3. Group decomposition: Q1→Q3 adjustments will significantly reduce the residual gap in ERT between dyslexics and controls (improving reading times of both, but dyslexics more).
4. Exploratory decomposition: Dyslexics will show attenuated skipping rate effects across all three features while showing amplified TRT effects.

# Method

## Participants

Data from fifty-seven adults was drawn from the open-access CopCo dataset (38 controls, 19 dyslexic) (Hollenstein et al., 2022). The overall mean age was 31.88 years (SD = 10.98; range = 20–64). By group, controls averaged 29.79 years (SD = 7.92; *n* = 38) and dyslexic readers averaged 36.05 years (SD = 14.77; *n* = 19). The sample included 40 women and 17 men (controls: 28 F/10 M; dyslexic: 12 F/7 M). Most participants reported Danish as their native language (overall *n* = 44; controls *n* = 25; all dyslexic participants reported Danish). Reading/comprehension screening measures indicated mean comprehension accuracy of .80 (SD = .14) overall, with group means of .81 (controls) and .79 (dyslexic). Words-per-minute averaged 210.77 (SD = 74.18) overall, with 240.30 (SD = 60.74) for controls and 151.72 (SD = 63.36) for dyslexic readers.

## Materials and Task

Participants read multiple speeches/passages, with controls reading a mean of 4.94 speeches (SD = 1.94) and dyslexics 2.89 speeches (SD = 1.05), yielding word-by-word eye-movement data with identifiers for trial, paragraph, sentence, word position, and standard fixation-time measures (e.g., total reading time, gaze duration) plus skipping flags. In total, all participants read an average of 5799 words (SD=2505) with controls reading a mean of 6567 words (SD = 2500) while dyslexics read a mean of 4264 words (SD=1715) during data collection.

## Measures and Outcome Definitions

We analysed (a) skipping probability, (b) fixation duration when a word was fixated (total reading time; TRT), and (c) expected reading time (ERT). At the token level, ERT was defined as 0 ms for skipped words and equal to TRT for fixated words; model-based ERT predictions were joined from separate skip and duration models using our ERT equation: $\text{ERT} = (1 - P(\text{skip})) \times \text{TRT}_{\text{fixated}}$. This decomposition allowed pathway-specific inference for skipping vs. duration as well as a prediction of the overall effect of a feature on reading times via ERT.

## Data Preprocessing

***Data Formatting.*** Feature columns were standardized to length (in characters), frequency (in Zipf), surprisal (in bits), TRT (ms), and skip (0 or 1, with 1 = skipped). Extreme fixation-duration outliers (> 3 SD above the fixated mean) were removed. Pooled quartiles (Q1/Q3) were computed once for length, Zipf frequency, and surprisal, and "orientation checks" verified that Q1→Q3 corresponded to shorter → longer, rarer → more frequent, and more predictable → more surprising outputs, respectively. To address the strong length–frequency correlation ($r = -.80$) without residualization, we created seven equal-frequency pooled length bins once on the combined sample and re-used these bins and their pooled weights in all Zipf-conditional analyses of frequency.

***Lexical frequency (Zipf) Computation.*** We aggregated multiple 1-million-token Danish subcorpora from the Leipzig Corpora Collection (news, web, news crawl, Wikipedia, mixed sources). For each archive, lists were lower-cased and duplicate frequencies merged; tokens with count < 2 and punctuation-only strings were removed; up to the top 1.5-million-word types were retained which yielded 98.6% coverage with our dataset. Missing word frequencies (1.4%) were removed during analysis.

***Contextual Predictability (Surprisal) Computation.*** We estimated word-level surprisal using a causal Danish GPT-2 model (KennethTM/gpt2-medium-danish) scored strictly left-to-right, to mimic reading. Per-token negative log-likelihoods were aligned to words and summed; outputted nats were converted to bits (÷ ln 2).

## Statistical Modelling

We fit group-specific generalized additive models (GAMs) with additive smooths with no interaction terms included. The skip model was a binomial Logistic GAM predicting probability of skipping —*P*(skip)—with smooth terms for length, Zipf (monotonic increasing constraint), and surprisal. The duration model was a Linear GAM on log-TRT with smooths for length, Zipf (monotonic decreasing constraint), and surprisal; predictions were back-transformed to milliseconds using a smearing correction. Hyperparameters (number of splines, λ) were selected separately by group using a two-stage, subject-wise procedure: (1) pre-selection on a subject subsample via GroupKFold with grid-searched λ, applying the 1-SE rule to choose the simplest model within one SE of the best; (2) validation with 10-fold GroupKFold using the frozen hyperparameters (reporting AUC for skip and RMSE for duration). Final models were refit on all data with those frozen settings. We retained access to the skip and duration components for exploratory, pathway-specific analyses.

**Hypotheses and Inference**

***Hypothesis 1 – Feature Effects.*** For each feature, we defined the effect as the average marginal increase in ERT (in ms) between Q1 and Q3. For length and surprisal, we set only that feature to Q1 vs. Q3 while holding the other predictors at their group means. For frequency (Zipf), we evaluated effects conditionally within pooled length bins: within each bin, we set Zipf to that bin's Q1 vs. Q3, averaged per subject within bin, and then took the weighted average across bins using the pooled bin weights. We computed pathway-specific Q1 → Q3 shifts for skip, duration, and ERT.

***Hypothesis 2 – Dyslexic Amplification.*** Amplification was defined as a slope ratio (SR) for the Q3–Q1 contrast, computed per pathway (skip, duration, ERT):

$$\mathrm{SR} = \frac{|\Delta_{\mathrm{dys}}|}{|\Delta_{\mathrm{ctrl}}|}$$

Zipf SRs were computed within each pooled length bin and then averaged, by bin weight, to produce the final SR. Word length and surprisal were each computed across the entire group sample.

***Hypothesis 3 – Gap Decomposition.*** We defined a single canonical gap (G_0) as the subject-balanced mean difference in model-predicted ERT (dyslexic − control). The equal-ease counterfactual clamped text difficulty for both groups by setting length and surprisal to pooled Q1 (downward clamp) and Zipf to Q3 conditionally within pooled length bins. We then recomputed the between-group gap; the reduction from (G_0) quantified how much of the gap was explained by these text features. Feature-wise contributions to the ERT gap reduction were averaged across the 8 model configurations ($2^3$) obtained by toggling the three two-way interactions—length×frequency, length×predictability, and frequency×predictability—on/off (the three-way interaction was excluded). For our exploratory analysis, skipping vs. duration contributions were attributed via a two-order Shapley decomposition: we computed marginal contributions with skip added before duration and with duration added before skip, then averaged the two orders.

***Resampling and Decision Rules.*** All inferential uncertainty used subject-level bootstrapping (resampling subjects with replacement) with $n$ = 3,000 resamples. We report 95% percentile confidence intervals and two-tailed $p$ values using the conservative ((r+1)/(B+1)) correction; for hypothesis 2 we tested SRs against 1.0 (1.0 = no difference in slope between dyslexics and controls). Pathway-specific effects (skip, duration, ERT) are reported where relevant.

***Software.*** All analyses were implemented in Python using pyGAM for GAM fitting with GroupKFold cross-validation and custom utilities for caching, smearing correction, and subject-level bootstrapping; steps and resources for reproducing our analysis are provided in the open repository's README script.

## Results

### Preliminary Analyses

The analytic sample comprised 57 participants (19 dyslexic, 38 control) contributing 322,776-word tokens with complete identifiers, after all outliers and missing values were excluded. Overall skipping rate was ~0.38 (controls ≈ .40; dyslexic ≈ .28). Fixated-word total reading time (TRT) averaged 335 ms in controls and 486 ms in dyslexic readers, consistent with slower reading rates and lower skipping rates found in dyslexia. These descriptive patterns are reported here to contextualize the model-based results reported below.

### Model Outputs

Figure 1 shows fitted GAM smooths for controls (solid) and dyslexic readers (dashed) with 95% CIs; curves are trimmed to the 1st–99th percentiles. Figure 1 Panels A–C show predicted skipping probability P(skip); Panels D–F show total reading time when fixated (TRT); Panels G–I show expected reading time outcomes (ERT). Vertical dashed lines for each smooth mark the 25th and 75th percentiles of our predictor (used in Q1 → Q3 shifts). Model diagnostics are in Appendix Tables 1–2 and show our model performance to be comparable with previous literature. Our duration GAMs (controls $R^2$ = .089; dyslexic $R^2$ = .066) are comparable in magnitude to classic participant-level regressions that used the same three predictors (Kliegl et al., 2004; median $R^2$ ≈ .07).

**Figure 1**

*GAM smooths for length, frequency, and surprisal on skipping, TRT, and ERT.*

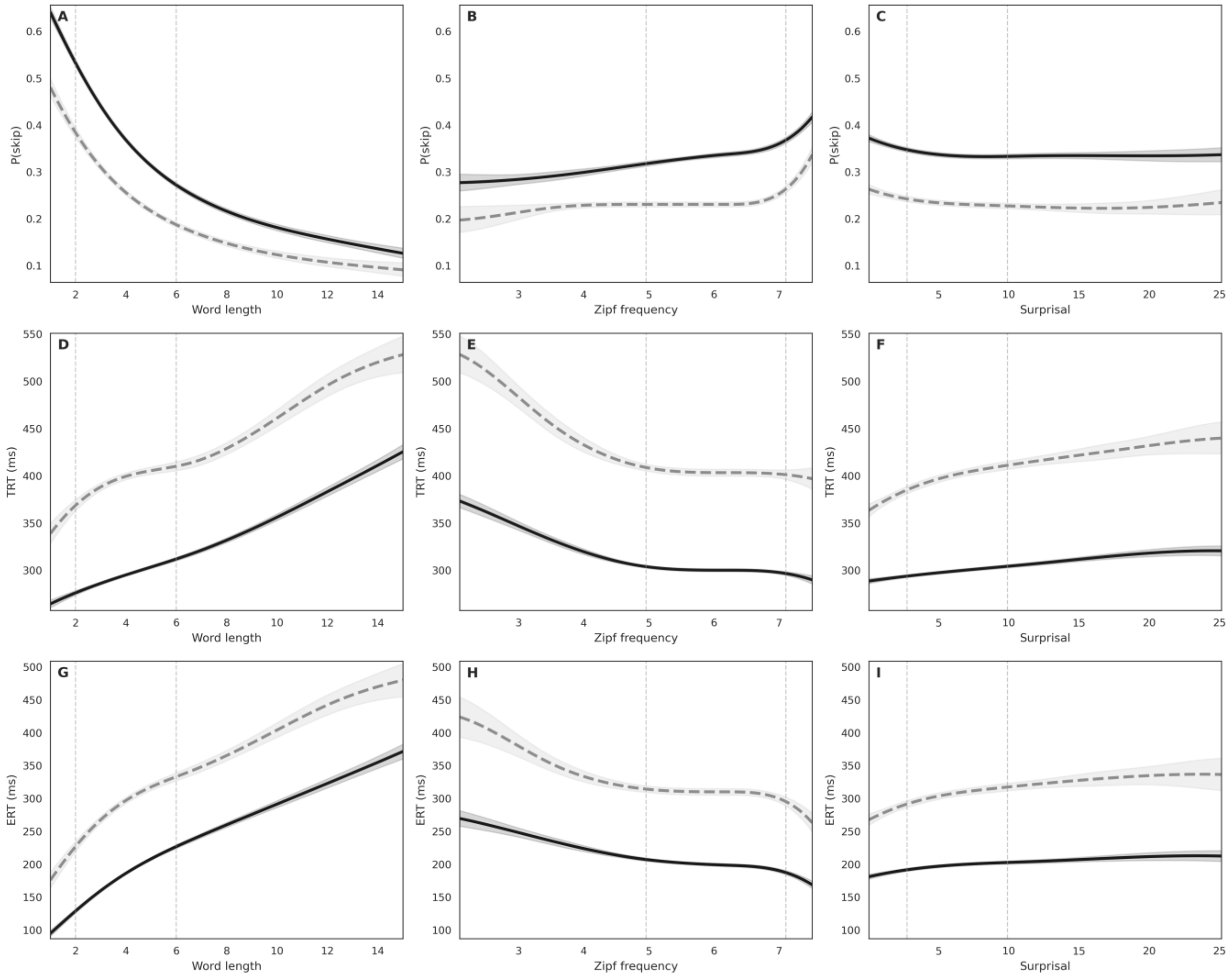

*Note.* Smooths are from group-specific GAMs with non-focal predictors held at the group means. Solid black line = Control; dashed grey line = Dyslexic; shaded bands = 95% CIs.

## Hypothesis 1: Feature Effects (Q1 → Q3)

Figure 2 shows the millisecond Q1→Q3 change in ERT for each feature and group. Length increased ERT in both groups, with a larger shift for dyslexic readers (controls: 98.99 ms, 95% CI [98.88, 99.11]; dyslexic: 108.87 ms, 95% CI [108.67, 109.09]; both $p < .001$). Zipf frequency (evaluated conditionally within length bins) reduced ERT in both groups, again with a larger reduction in dyslexia (controls: −17.22 ms, 95% CI [−17.50, −16.92]; dyslexic: −25.66 ms, 95% CI [−26.83, −24.53]). Surprisal increased ERT in both groups, with a substantially larger effect for dyslexic readers (controls: 10.65 ms, 95% CI [10.61, 10.69]; dyslexic: 24.98 ms, 95% CI [24.85, 25.12]; all $p < .001$).

**Figure 2**

*Feature effects Q1 → Q3 on ERT with 95% CIs.*

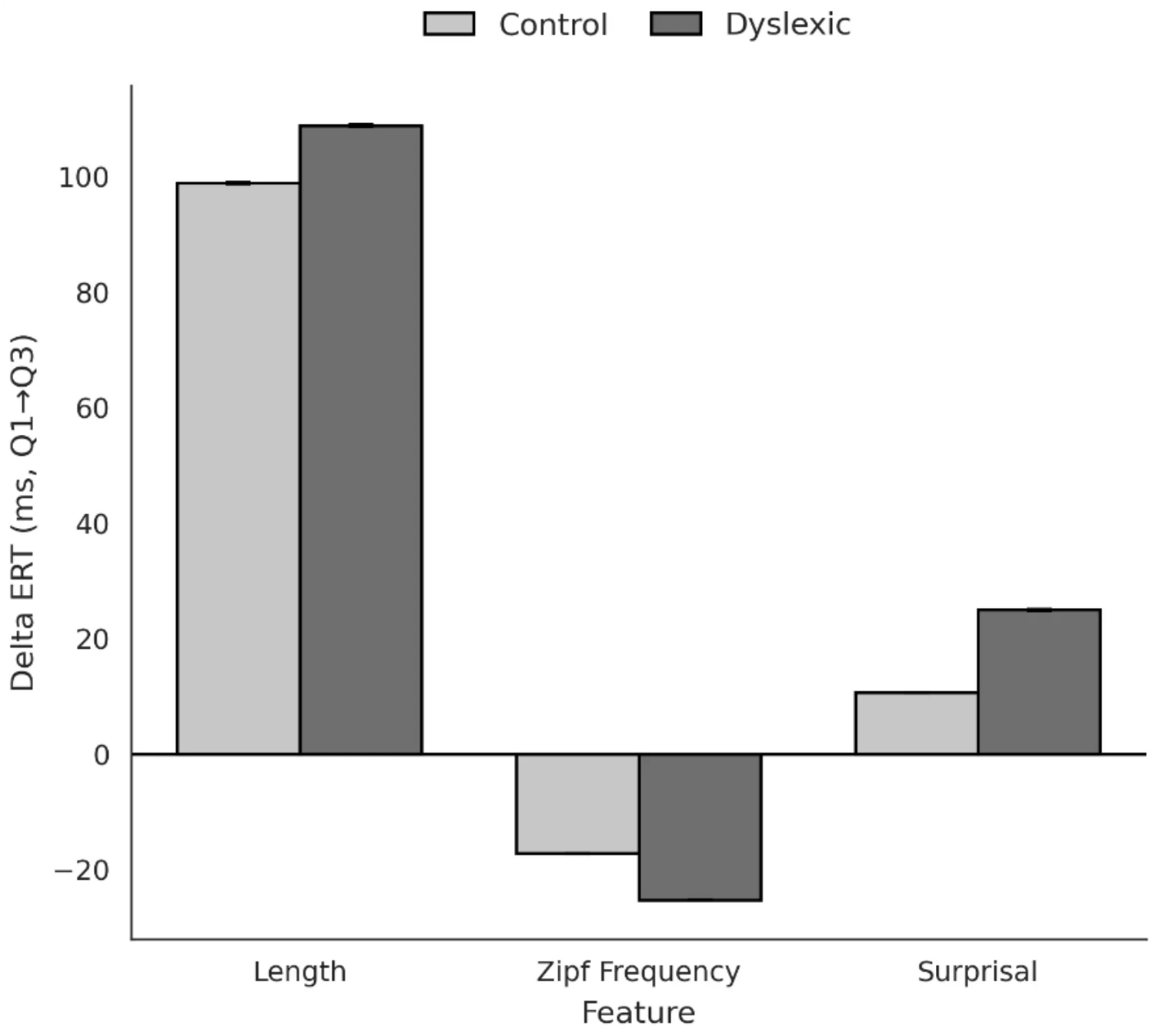

*Note.* Zipf Q1→Q3 is evaluated conditionally within length bins to account for collinearity. Zipf is also negative as moving from Q1→Q3 is a high→low frequency change.

**Hypothesis 2: Dyslexic Amplification**

Figure 3 summarizes slope ratios (*SR*s) for ERT and by pathway. Length showed modest overall amplification on ERT ($SR = 1.08$, 95% CI [1.083, 1.083], $p < .001$), driven by amplification in the duration pathway ($SR = 1.16$, 95% CI [1.155, 1.156], $p < .001$) and attenuation in skipping ($SR = 0.75$, 95% CI [0.755, 0.755], $p < .001$). Zipf showed clear overall amplification on ERT ($SR = 1.34$, 95% CI [1.29, 1.39], $p < .001$), with attenuation in skipping ($SR = 0.80$, 95% CI [0.79, 0.83], $p < .001$) and modest amplification in duration ($SR = 1.08$, 95% CI [1.00, 1.16], $p = .047$). Surprisal yielded the strongest amplification: ERT $SR = 2.32$ (95% CI [2.316, 2.326], $p < .001$), with $SR = 2.50$ for duration (95% CI [2.480, 2.519], $p < .001$) and a small amplification in skipping ($SR = 1.04$, 95% CI [1.03, 1.06], $p < .001$).

**Figure 3**

*Dyslexic slope ratio (SR) amplification by pathway with 95% CIs.*

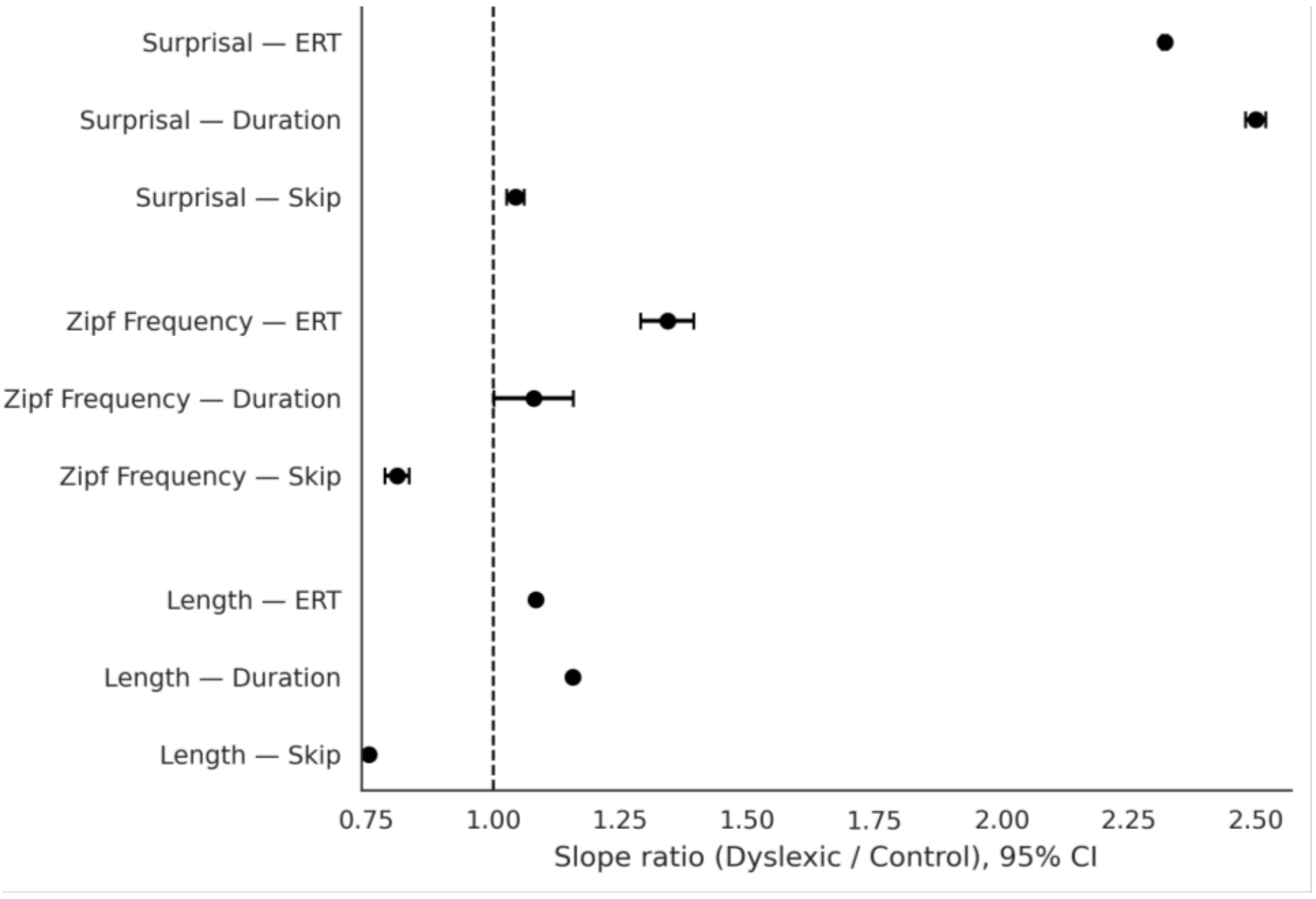

*Note.* Vertical line marks SR = 1 (parity). Estimates use word-level GAMs fit separately by group (binomial for skipping; log-Gaussian for duration with smearing back-transform).

**Hypothesis 3: Decomposition of the Dyslexic–Control Gap**

Figure 4 reports three decomposition analyses. The between-group ERT gap was 97.28 ms/word (95% CI [94.25, 100.51], $p < .001$). Shapley decomposition attributed 63.21 ms/word (95% CI [62.53, 63.89]) to the duration pathway and 34.07 ms/word (95% CI [33.89, 34.26]) to reduced skipping—approximately 65% and 35% of the gap, respectively, shown in Figure 4B. Under an equal-ease counterfactual—clamping length and surprisal to easier (Q1) values and Zipf to higher, bin-wise Q3 values in both groups—the gap shrank by 30.66 ms/word (95% CI [27.73, 33.69]), leaving a counterfactual gap of 66.61 ms/word (95% CI [66.32, 66.85]; both $p < .001$), shown in Figure 4A. Feature-wise contributions to this reduction were: surprisal, 13.67 ms/word (95% CI [12.88, 14.46]); length, 9.98 ms/word (95% CI [6.49, 13.37]); and Zipf, 7.01 ms/word (95% CI [6.15, 7.90]). Taken together, simplifying lexical difficulty (shorter, more frequent, more predictable words) eliminated approximately 32% of the observed gap, shown in Figure 4C.

**Figure 4**

*Decomposition of the dyslexic–control expected reading-time (ERT) gap with 95% CIs.*

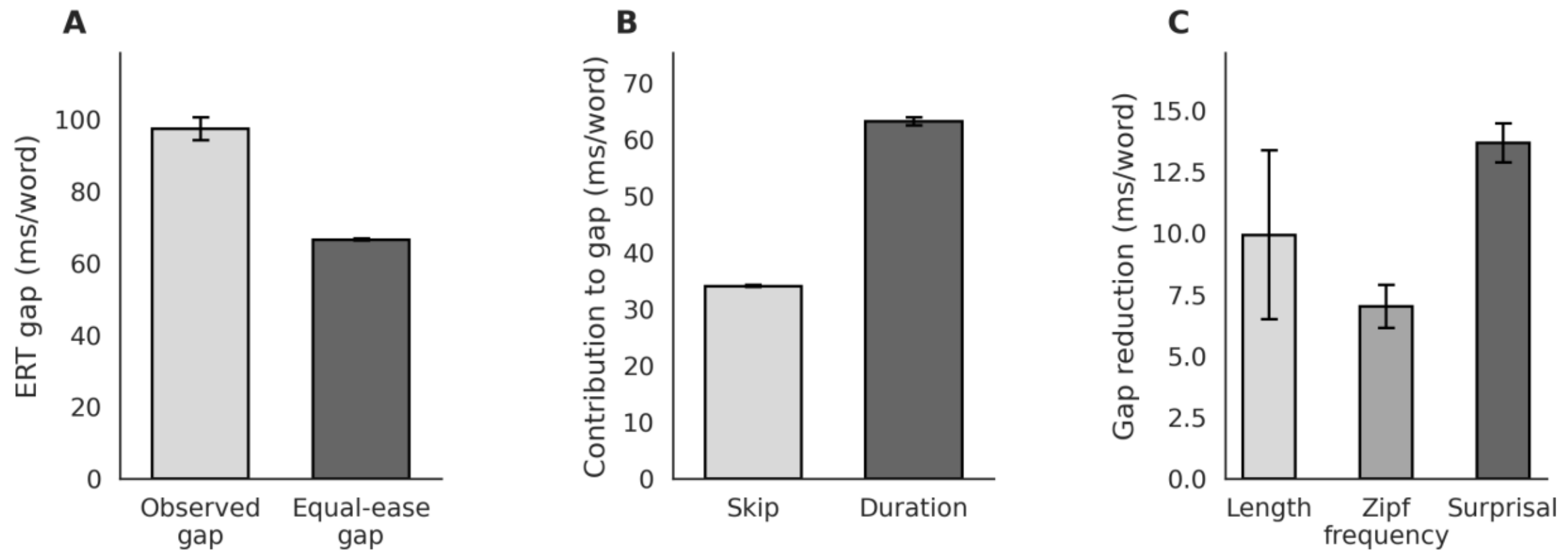

*Note.* Panel A: Total between-group ERT gap and final reduced gap following equal-ease counterfactual. Panel B: Skip and duration bars sum to the observed gap. Panel C: Feature-wise contributions to the change in the counterfactual gap.

## Discussion

Across naturalistic connected-text reading, three word-level features—length, lexical frequency (Zipf), and predictability (surprisal)—showed strong, directionally predicted effects on expected reading time (ERT). Dyslexic readers exhibited larger ERT sensitivities than controls for all three features (length SR≈1.08; Zipf SR≈1.34; surprisal SR≈2.32; all *ps* < .001), directly supporting H1 and H2 and showing that findings already seen in typical readers (Kliegl et al., 2004; Rayner et al., 2011) extend to dyslexics. In single-feature contrasts, word length shifts reduced the dyslexic–control gap by ≈9.98 ms/word, frequency reduced it by ≈7.01 ms/word, and surprisal by ≈13.67 ms/word (all *ps* < .001). When all three features were changed together (from the 25th → 75th percentile), the dyslexic–control gap cumulatively decreased from ≈97.28 ms/word to ≈66.61 ms/word (a reduction of 30.66 ms or ≈32% of the total dyslexic gap; *p*<.001). These consistent reductions support H3, showing that the three word-level features we tested each produced large, disproportionate gap reductions for dyslexic readers versus controls.

Beyond replication and extension of existing evidence, surprisal emerged as the strongest (new) dyslexia-relevant feature: it showed, by far, the largest ERT amplification (SR≈2.32) and the largest single-feature gap reduction (13.67ms), indicating a uniquely high sensitivity among dyslexic readers to contextual predictability. From the lens of the dyslexia theories we presented, the surprisal effect strongly affirms the constrained working memory account of dyslexia (Maehler & Schuchardt, 2016; Du & Zhang, 2023; Smith-Spark & Fisk, 2007), which predict that words burdening linguistic working memory should disproportionately affect reading in dyslexics (Engelhardt et al., 2021; Hawelka et al., 2015). Since word surprisal is just one of a multitude features affecting working memory at the word level, these findings encourage further exploration into how much of the remaining dyslexic reading gap these features may collectively explain (discussed in future directions). Additionally, in line with the phonological account of dyslexia (Ziegler & Goswami, 2005; Hyönä & Olson, 1995), longer and rarer words were also anticipated to disproportionately affect reading times by slowing lexical access and phonemic conversion. Although these effects were smaller in magnitude, there were clear ERT amplifications in both length and frequency, supporting the phonological account as well. Taken as a whole, our findings demonstrate that working memory differences had a stronger reading-time influence on dyslexics although phonological effects also had a fairly strong, smaller influence.

Our exploratory, pathway decomposition localized two thirds of the total dyslexic gap to the fixation duration pathway (63.20 ms), with skipping accounting for a smaller, one-third of the gap (34.07 ms). Additionally, our skip-pathway slope ratios showed strong *attenuations* for length (≈0.75) and Zipf (≈0.80) in dyslexics, and very small amplifications for surprisal (≈1.04). These patterns fit with existing findings which consistently show reductions in skipping amongst dyslexics (Hutzler & Wimmer, 2004; Hawelka et al., 2010) and suggest that our features overwhelmingly improved reading times through reductions in fixation times, not skipping, supporting our exploratory hypothesis. This also implies manipulations in word length, frequency, and surprisal are mostly ineffective at increasing rates of skipping for dyslexics; thus, other factors should be considered to fill this gap (explored in future directions).

We addressed several possible confounds over the course of this study. First, we observed a strong length–frequency collinearity ($r = -.80$) which we accounted for by evaluating frequency conditionally within weighted length bins to estimate ms-shifts. Since length and surprisal showed substantially weaker collinearities, we took the standard approach of varying the target feature while holding other features at group means for our ms-shift estimates. Our predictive modelling strategy included all three predictors simultaneously as additive smooths, to mitigate confounding. However, interactions (e.g., length×surprisal; length×frequency) were not explicitly included in our final model. We initially considered interactions but found these showed no improvements on held-out performance, distorting smooths, while significantly increasing the complexity of our model; additionally, prior research found very little or no interactions between either length, predictability, or frequency (Rayner et al., 2004, 2011; Kliegl et al., 2004), suggesting little benefit to this approach. Therefore, to improve the generalizability and reduce likely overfitting we decided to simplify our modelling approach to just include three main effects: length, frequency, and surprisal. This was successful and maintained performance on each of the key metrics cited. Finally, we used conservative percentile contrasts (25th↔75th percentile shifts) to limit extrapolation, ensuring results were likely to generalize across texts, languages, and experimental conditions, while avoiding extreme estimates. This likewise resulted in clear, usable numbers.

Despite these considerations, an intrinsic limitation to our study was the purely observational data which our model and gap-reduction statistics depended on. This means our modelling choices (e.g. how we constrained our model and which predictors we omitted) could have biased the millisecond contrasts we presented. Further studies could account for this possibility by extending our model to a new dataset (new language, texts, and features) or—even more usefully—implementing an equivalent, randomized manipulation within a naturalistic reading scenario to compare outcomes with our model's ERT predictions. Attempting a preliminary evaluation against the literature, when we compared our modelled control outputs against the closest equivalent study—the findings of Kliegl and colleagues (2004)—we found effect sizes and directions to be mutually consistent with our model. The order of magnitudes and direction of values (see our figure 2) was the same—length (largest) → frequency → predictability (smallest)—despite the study being conducted in German and on a much smaller word corpus (1.1K vs 245.1K words). Our ERT slope curves likewise showed qualitatively similar shapes and landmarks to the duration and skipping functions reported by Kliegl and colleagues (2004): with a monotonic rise with length, approximately log-linear frequency effects, and a comparatively smaller predictability influence on durations (see figure 1.A-F). This offers preliminary support for our modelling choices. When compared against the literature, our models also showed more conservative estimates than other studies (e.g. Rayner et al., 2011; Kliegl et al., 2004); this potential underestimation is expected, and most likely reflects the purposefully naturalistic (vs. highly controlled) setting of our study, as well as the choice to employ a smaller (25→75th percentile) shift for estimating main effects. Unfortunately, we could find no large-scale dyslexia studies which used a similar methodology to ours, so we could not make an equivalent comparison for our dyslexic outputs.

Considering future directions, despite clear positive effects, a sizeable residual gap remains (≈66.61 ms/word; ≈68% of the original gap). This means our three word-level features—length, frequency, surprisal—are important but not sufficient to close the reading gap, leaving space for additional features to be considered. Based on our findings, we suggest two novel opportunities to explain additional portions of the dyslexic gap:

First, because surprisal showed the strongest dyslexic amplification (consistent with working memory constraint theories, which imply weakened word prediction), it may be beneficial to examine additional working-memory-relevant word features for dyslexics. In this area, syntactic complexity appears the most promising to examine. Research consistently shows that syntactic complexity loads on linguistic working memory in typical readers (Gibson, 1998), and large eye-tracking corpora show that its measurement can predict word-by-word reading times (Boston et al., 2008; Demberg & Keller, 2008). While there are many methods for measuring linguistic complexity, the strongest per-word candidates are: (1) DLT storage cost—the number of open dependencies maintained at each word (Gibson, 1998, 2000); (2) DLT integration cost—the distance-based cost incurred when a dependency closes at a word (Gibson, 2000; Grodner & Gibson, 2005); and (3) parser stack/embedding depth (e.g., Yngve or left-corner stack depth) as a direct working-memory load proxy at each word (Yngve, 1960; Resnik, 1992). An exploration of either (or all) of these within our models, could reasonably build on our findings that linguistic working memory can explain a large portion of the observed dyslexic time gap.

Second, our exploratory decomposition showed that skipping accounts for roughly one-third of the total reading gap, despite being largely unchanged by our three, word-level manipulations (skip-path SRs: length ≈0.75; Zipf ≈0.80; surprisal ≈1.04). This leaves ample room for dyslexia-specific factors, which target the skipping pathway, to fill this gap. One strong candidate in this area is the (spatially driven) parafoveal preview benefit—the likelihood of extracting information from upcoming letters and words before fixation. Dyslexic readers consistently show reduced preview benefits in eye-tracking tasks (Silva et al., 2016; Yan et al., 2013), and these reductions chiefly decrease skip probability (Angele et al., 2014; Gordon et al., 2013; Choi & Gordon, 2014), making it particularly relevant. Thus, if preview benefit were modelled alongside our other features, it could directly target the skipping pathway (which our study has mostly failed to change in dyslexics).

In sum, length, frequency, and surprisal levels change reading times in both groups, but dyslexic readers are more sensitive to all three, especially surprisal. Simplifying these features reduces the dyslexic–control time gap by ~one-third, with the remaining gap residing in the fixation duration pathway. Our clear, significant results provide actionable, word-level factors for designing new interventions while motivating next steps to examine additional predictors and experimental validations to improve beyond our current model. Together, we offer the first answer to why dyslexic reading takes longer—and when, supporting new reading models and interventions built with our findings.

**Reproducibility Statement**

Under the CC BY-NC-SA license, you are welcome to copy and reuse all code and models use in the development of this paper. All of our resources are stored under: https://github.com/hugorydel/Dyslexia_Time_Cost_Analysis

**References**

Angele, B., Laishley, A. E., Rayner, K., & Liversedge, S. P. (2014). The effect of high- and low-frequency previews and sentential fit on word skipping during reading. *Journal of Experimental Psychology: Learning, Memory, and Cognition, 40*(4), 1181–1203. https://doi.org/10.1037/a0036396

Boston, M. F., Hale, J. T., Kliegl, R., Patil, U., & Vasishth, S. (2008). Parsing costs as predictors of reading difficulty: An evaluation using the Potsdam Sentence Corpus. *Journal of Eye Movement Research, 2*(1), 1–12. https://doi.org/10.16910/jemr.2.1.1

Choi, W., & Gordon, P. C. (2014). Word skipping during sentence reading: Effects of lexicality on parafoveal processing. *Attention, Perception, & Psychophysics, 76*(1), 201–213. https://doi.org/10.3758/s13414-013-0494-1

Clifton, C., Jr., Staub, A., & Rayner, K. (2007). Eye movements in reading words and sentences. In R. P. G. van Gompel, M. H. Fischer, W. S. Murray, & R. L. Hill (Eds.), *Eye movements: A window on mind and brain* (pp. 341–372). Elsevier. https://doi.org/10.1016/B978-008044980-7/50017-3

De Luca, M., Borrelli, M., Judica, A., Spinelli, D., & Zoccolotti, P. (2002). Reading words and pseudowords: An eye movement study of developmental dyslexia. *Brain and Language, 80*(3), 617–626. https://doi.org/10.1006/brln.2001.2637

Demberg, V., & Keller, F. (2008). Data from eye-tracking corpora as evidence for theories of syntactic processing complexity. *Cognition, 109*(2), 193–210. https://doi.org/10.1016/j.cognition.2008.07.008

Doust, C., Fontanillas, P., Eising, E., Ebejer, J. L., Webster, M. J., Wang, Z., St Pourcain, B., Schumacher, J., Talcott, J. B., Ystrom, E., Willcutt, E. G., DeFries, J. C., Olson, R. K., Pennington, B. F., Smith, S. D., Gayán, J., Geschwind, D. H., Kere, J., … Paracchini, S. (2022). Discovery of 42 genome-wide significant loci associated with

dyslexia. *Nature Genetics, 54*(11), 1621–1629. https://doi.org/10.1038/s41588-022-01192-y

Du, Y., & Zhang, C. (2023). The deficit in visuo-spatial working memory in dyslexic population? A systematic review and meta-analysis. *International Journal on Social and Education Sciences, 5*(2), 295–306. https://doi.org/10.46328/ijonses.506

Engbert, R., Nuthmann, A., Richter, E. M., & Kliegl, R. (2005). SWIFT: A dynamical model of saccade generation during reading. *Psychological Review, 112*(4), 777–813. https://doi.org/10.1037/0033-295X.112.4.777

Engelhardt, P. E., Yuen, M. K. Y., Kenning, E. A., & Filipović, L. (2021). Are linguistic prediction deficits characteristic of adults with dyslexia? *Brain Sciences, 11*(1), 59. https://doi.org/10.3390/brainsci11010059

Gibson, E. (1998). Linguistic complexity: Locality of syntactic dependencies. *Cognition, 68*(1), 1–76. https://doi.org/10.1016/S0010-0277(98)00034-1

Gibson, E. (2000). Dependency locality theory: A distance-based theory of linguistic complexity. In A. Marantz, Y. Miyashita, & W. O'Neil (Eds.), *Image, language, brain: Papers from the first Mind Articulation Project symposium* (pp. 95–126). MIT Press.

Gordon, P. C., Plummer, P., & Choi, W. (2013). See before you jump: Full recognition of parafoveal words precedes skips during reading. *Journal of Experimental Psychology: Learning, Memory, and Cognition, 39*(2), 633–641. https://doi.org/10.1037/a0028881

Grodner, D., & Gibson, E. (2005). Consequences of the serial nature of linguistic input for sentential complexity. *Cognitive Science, 29*(2), 261–290. https://doi.org/10.1207/s15516709cog0000_7

Hawelka, S., Gagl, B., & Wimmer, H. (2010). A dual-route perspective on eye movements of dyslexic readers. *Cognition, 115*(3), 367–379. https://doi.org/10.1016/j.cognition.2009.11.004

Hawelka, S., Schuster, S., Gagl, B., Hutzler, F., & Richlan, F. (2015). On forward inferences of fast and slow readers: An eye movement study. *Scientific Reports, 5*, 8432. https://doi.org/10.1038/srep08432

Hollenstein, N., Björnsdóttir, M., & Barrett, M. (2022). *CopCo: The Copenhagen Corpus of eye-tracking recordings from natural reading* [Data set]. CLARIN-DK-UCPH Repository. http://hdl.handle.net/20.500.12115/48

Hyönä, J., & Olson, R. K. (1995). Eye fixation patterns among dyslexic and normal readers: Effects of word length and word frequency. *Journal of Experimental Psychology: Learning, Memory, and Cognition, 21*(6), 1430–1440. https://doi.org/10.1037/0278-7393.21.6.1430

Hutzler, F., & Wimmer, H. (2004). Eye movements of dyslexic children when reading in a regular orthography. *Brain and Language, 89*(2), 235–242. https://doi.org/10.1016/S0093-934X(03)00401-2

Juhasz, B. J., White, S. J., Liversedge, S. P., & Rayner, K. (2008). Eye movements and the use of parafoveal word length information in reading. *Journal of Experimental Psychology: Human Perception and Performance, 34*(6), 1560–1579. https://doi.org/10.1037/a0012319

Kliegl, R., Grabner, E., Rolfs, M., & Engbert, R. (2004). Length, frequency, and predictability effects of words on eye movements in reading. *European Journal of Cognitive Psychology, 16*(1–2), 262–284. https://doi.org/10.1080/09541440340000213

Kutas, M., & Federmeier, K. D. (2011). Thirty years and counting: Finding meaning in the N400 component of the event-related brain potential (ERP). *Annual Review of Psychology, 62*, 621–647. https://doi.org/10.1146/annurev.psych.093008.131123

Lyon, G. R., Shaywitz, S. E., & Shaywitz, B. A. (2003). A definition of dyslexia. *Annals of Dyslexia, 53*(1), 1–14. https://doi.org/10.1007/s11881-003-0001-9

Maehler, C., & Schuchardt, K. (2016). Working memory in children with specific learning disorders and/or attention deficits. *Learning and Individual Differences, 49*, 341–347. https://doi.org/10.1016/j.lindif.2016.05.007

Pelli, D. G., & Tillman, K. A. (2008). The uncrowded window of object recognition. *Nature Neuroscience, 11*(10), 1129–1135. https://doi.org/10.1038/nn.2187

Perfetti, C. A. (2007). Reading ability: Lexical quality to comprehension. *Scientific Studies of Reading, 11*(4), 357–383. https://doi.org/10.1080/10888430701530730

Peterson, R. L., & Pennington, B. F. (2015). Developmental dyslexia. *Annual Review of Clinical Psychology, 11*, 283–307. https://doi.org/10.1146/annurev-clinpsy-032814-112842

Pollatsek, A., Juhasz, B. J., Reichle, E. D., Machacek, D., & Rayner, K. (2008). Immediate and delayed effects of word frequency and word length on eye movements in reading: A reversed delayed effect of word length. *Journal of Experimental Psychology: Human Perception and Performance, 34*(3), 726–750. https://doi.org/10.1037/0096-1523.34.3.726

Rayner, K. (1998). Eye movements in reading and information processing: 20 years of research. *Psychological Bulletin, 124*(3), 372–422. https://doi.org/10.1037/0033-2909.124.3.372

Rayner, K., & Well, A. D. (1996). Effects of contextual constraint on eye movements in reading: A further examination. *Psychonomic Bulletin & Review, 3*(4), 504–509. https://doi.org/10.3758/BF03214555

Rayner, K., Ashby, J., Pollatsek, A., & Reichle, E. D. (2004). The effects of frequency and predictability on eye fixations in reading: Implications for the E-Z Reader model. *Journal of Experimental Psychology: Human Perception and Performance, 30*(4), 720–732. https://doi.org/10.1037/0096-1523.30.4.720

Rayner, K., Slattery, T. J., Drieghe, D., & Liversedge, S. P. (2011). Eye movements and word skipping during reading: Effects of word length and predictability. *Journal of Experimental Psychology: Human Perception and Performance, 37*(2), 514–528. https://doi.org/10.1037/a0020990

Reichle, E. D., Liversedge, S. P., Drieghe, D., Blythe, H. I., Joseph, H. S., White, S. J., & Rayner, K. (2013). Using E-Z Reader to examine the concurrent development of eye-movement control and reading skill. *Developmental review : DR*, *33*(2), 110–149. https://doi.org/10.1016/j.dr.2013.03.001

Reichle, E. D., Rayner, K., & Pollatsek, A. (2003). The E-Z Reader model of eye-movement control in reading: Comparisons to other models. *Behavioral and Brain Sciences, 26*(4), 445–476. https://doi.org/10.1017/S0140525X03000104

Reichle, E. D., Warren, T., & McConnell, K. (2009). Using E-Z Reader to model the effects of higher-level language processing on eye movements during reading. *Psychonomic Bulletin & Review, 16*(1), 1–21. https://doi.org/10.3758/PBR.16.1.1

Resnik, P. (1992). Left-corner parsing and psychological plausibility. In *Proceedings of the 14th International Conference on Computational Linguistics (COLING-1992)* (pp. 191–197). Association for Computational Linguistics. https://doi.org/10.3115/992066.992098

Schotter, E. R., & Leinenger, M. (2016). Reversed preview benefit effects: Forced fixations emphasize the importance of parafoveal vision for efficient reading. *Journal of Experimental Psychology: Human Perception and Performance, 42*(12), 2039–2067. https://doi.org/10.1037/xhp0000270

Schotter, E. R., Angele, B., & Rayner, K. (2012). Parafoveal processing in reading. *Attention, Perception, & Psychophysics, 74*(1), 5–35. https://doi.org/10.3758/s13414-011-0219-2

Silva, S., Faísca, L., Araújo, S., Casaca, L., Carvalho, L., Petersson, K. M., & Reis, A. (2016). Too little or too much? Parafoveal preview benefits and parafoveal load costs in dyslexic adults. *Annals of Dyslexia, 66*(2), 187–201. https://doi.org/10.1007/s11881-015-0113-z

Slattery, T. J., Ganesh, S., Martin, J. D., Spears, R., & Rayner, K. (2018). Word skipping: Effects of word length, predictability, spelling, and reading skill. *Quarterly Journal of Experimental Psychology, 71*(12), 2507–2518. https://doi.org/10.1080/17470218.2017.1310264

Smith-Spark, J. H., & Fisk, J. E. (2007). Working memory functioning in developmental dyslexia. *Memory, 15*(1), 34–56. https://doi.org/10.1080/09658210601043384

Staub, A. (2011). The effect of lexical predictability on distributions of eye fixation durations. *Psychonomic Bulletin & Review, 18*(2), 371–376. https://doi.org/10.3758/s13423-010-0046-9

Staub, A. (2015). The effect of lexical predictability on eye movements in reading: *A critical review and theoretical interpretation*. *Language and Linguistics Compass, 9*(8), 311–327. https://doi.org/10.1111/lnc3.12151

Tiffin-Richards, S. P., & Schroeder, S. (2015). Word length and frequency effects on children's eye movements during silent reading. *Vision Research, 113*(B), 33–43. https://doi.org/10.1016/j.visres.2015.05.008

Warren, T., McConnell, K., & Rayner, K. (2008). Effects of context on eye movements when reading about possible and impossible events. *Journal of Experimental Psychology: Learning, Memory, and Cognition, 34*(4), 1001–1010. https://doi.org/10.1037/0278-7393.34.4.1001

White, S. J., Rayner, K., & Liversedge, S. P. (2005). The influence of parafoveal word length and contextual constraint on fixation durations and word skipping in reading. *Psychonomic Bulletin & Review, 12*(3), 466–471. https://doi.org/10.3758/BF03193789

Yan, M., Pan, J., Laubrock, J., Kliegl, R., & Shu, H. (2013). Parafoveal processing efficiency in rapid automatized naming: A comparison between Chinese normal and dyslexic children. *Journal of Experimental Child Psychology, 115*(3), 579–589. https://doi.org/10.1016/j.jecp.2013.01.007

Yang, L., Li, C., Li, X., Zhai, M., An, Q., Zhang, Y., Zhao, J., & Weng, X. (2022). Prevalence of developmental dyslexia in primary school children: A systematic review and meta-analysis. *Brain Sciences, 12*(2), 240. https://doi.org/10.3390/brainsci12020240

Yngve, V. H. (1960). A model and an hypothesis for language structure. *Proceedings of the American Philosophical Society, 104*(5), 444–466.

Ziegler, J. C., & Goswami, U. (2005). Reading acquisition, developmental dyslexia, and skilled reading across languages: A psycholinguistic grain size theory. *Psychological Bulletin, 131*(1), 3–29. https://doi.org/10.1037/0033-2909.131.1.3

## Appendix

**Table 1**

*Diagnostics for Skipping Model (Binomial Logistic)*

| Metric | Controls | Dyslexics |
|---|---|---|
| *n* (observations) | 245,104 | 77,672 |
| AUC | 0.696 | 0.683 |
| Cross-validated AUC | 0.700 | 0.682 |

**Table 2**

*Diagnostics for Duration Model (Log-Scale Gaussian)*

| Metric | Controls | Dyslexics |
|---|---|---|
| *n* (observations) | 146,325 | 54,179 |
| $R^2$ | 0.089 | 0.066 |
| Cross-validated RMSE (log) | 0.521 | 0.577 |

*Note.* Duration models were fit on log-transformed fixation duration; RMSE is reported on the log scale. $R^2$ values are comparable to prior models on these features.